# Detection of Slang Words in e-Data using semi-Supervised Learning

Alok Ranjan Pal[1] and Diganta Saha[2]

[1]College of Engineering and Management, Kolaghat, West Bengal, India, 721171
[2]Jadavpur University, Kolkata, West Bengal, India, 700098

## *Abstract*

*The proposed algorithmic approach deals with finding the sense of a word in an electronic data. Now a day, in different communication mediums like internet, mobile services etc. people use few words, which are slang in nature. This approach detects those abusive words using supervised learning procedure. But in the real life scenario, the slang words are not used in complete word forms always. Most of the times, those words are used in different abbreviated forms like sounds alike forms, taboo morphemes etc. This proposed approach can detect those abbreviated forms also using semi supervised learning procedure. Using the synset and concept analysis of the text, the probability of a suspicious word to be a slang word is also evaluated.*

## 1. INTRODUCTION

The Internet technology and the Telecommunication system have become indispensable in everyone's life today. The obvious reason is people can share information with other around the world via e-mail, chatting, community forums, SMS etc. Through these electronic mediums people share different types of information related to study, research, business, sports, entertainment, national and international affairs etc.

But this technological revolution has some dark side also. While sharing the information, sometimes people use jargon words, which appear openly in different public community forums. This type of malfunctioning on web has some instant as well as long time effect on society. Few anti social communities use this medium with bad intension to spread rumor and violence. Recently, Governments of different countries are taking different steps to protect the malfunctioning over Internet and Telecommunication system, as the use of few particular words are restricted in SMS service, few selected sites are blocked and discussions on community forums are strictly monitored etc.

This algorithmic approach can detect the jargon words in different e-texts, if it is used as a script in a data handling procedure and the e-texts are passed through this script before submission on the web or any network.

Organization of rest of the paper is as follows: Section 2 is about motivation of our paper; Section 3 describes the background; Section 4 depicts the proposed approach in detail; Section 5 depicts experimental results; Section 6 represents conclusion of the paper.

## 2. MOTIVATION

Recently, to stop this malfunctioning Governments of different countries are taking different effective steps. In Pakistan, use of around 1700 specific words on any kind of communication medium is strictly prohibited. In USA, CHINA and few countries access of few web sites are restricted. Most recently, Government of India has warned few community sites to monitor the data shared through their systems.

This proposed algorithm can filter the abusive words as well as suspicious word forms in an e-text when those are shared on a medium

## 3. BACKGROUND

Natural Language Processing (NLP) [1] is one of the challenging tasks in recent day's research field. Many research works are carried out in different sub-domains of NLP, as Information Retrieval (IR), Automated Classification [7], Language Translation by Machine [4, 5, 6], Word Sense Disambiguation [9, 10, 17], Part of speech Tagging [15, 16], Anaphora Resolution [11], Paraphrasing [12], Malapropism [13], Collocation Testing [14] etc. These research works greatly depend on some knowledge driven methods [8]. WordNet [2, 18, 20] is a knowledge source, which is used greatly in different knowledge based research works. The key features of this knowledge base are, it is machine readable and in this knowledge base the words are arranged semantically instead of alphabetically. The words of symmetric sense create a group, called *synset* and each *synset* represents a particular sense, called *concept* [19].This semantic architecture of words play a vital role in different knowledge based research works of NLP like Automatic Summarisation, E-Learning [3], Automatic Medical Diagnosis etc.

Our approach adopts the idea of resolution of sense of a word from a given text.

## 4. PROPOSED APPROACH

This algorithmic approach handles the slang words in four ways. First, using supervised learning methodology the jargon words of complete word form are detected by the help of an initially populated jargon word Database (Figure 10).

An additional job is performed in this stage. Using the *synset* analysis, the context of the discussion (concept) [19] is evaluated.

Secondly, the jargon words, which are used in some abbreviated forms but sound like a jargon word, are detected by the help of a sounds-alike jargon word Database (Figure 11).

Thirdly, the algorithm deals with those words, which are suspicious in nature to be a jargon word (Figure 5). Using a sliding window mechanism the partially matched jargon words are detected and stored in a temporary Database with a counter value (Figure 13) for further verification.

Finally, the suspicious words are verified by the help of *synset* and *concept* analysis as they can be used for further decision making.

The proposed approach is described below using modular representation:

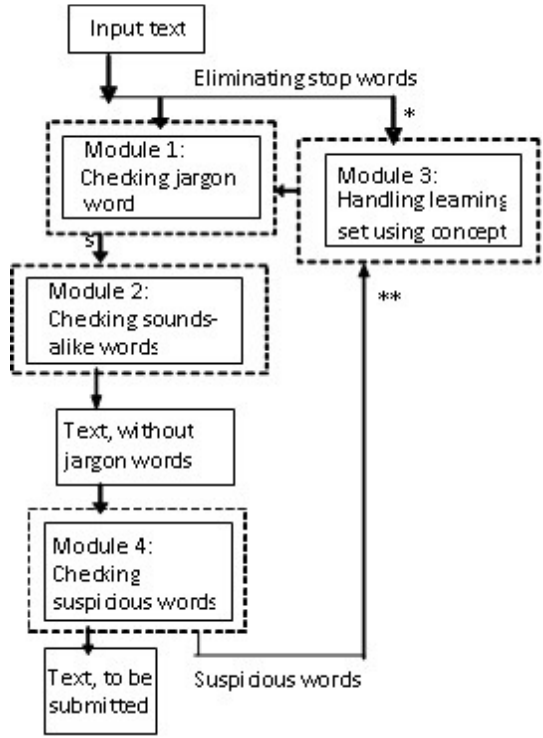

Figure 1. Modular representation of the overall Procedure

**Algorithm 1:** The over all procedure is described in Figure 1, where each module performs a particular task. In module 1, the barely appeared words are detected using supervised learning. In module 2, the sounds-alike abbreviated forms of slang words are detected. The suspicious words are detected in Module 4 and the probability analysis for a suspicious word to be a jargon word is handled in Module 3.

Input: Input text.
Output: Text to be submitted.

1. Stop words are eliminated from the input text.
2. Text, with only meaningful words, is created.
3. This text is passed to
   a. Module 1 for detecting the slang words.
   b. Module 3 for deriving the concept from the text.
4. Text, without completely matched slang words, is obtained from Module 1.
5. The text from Module 1 is passed to Module 2, where the sounds-alike slang words are checked.
6. The text from Module 2 is passed to Module 4, where the occurrence of any suspicious word is checked.

51

If any suspicious word is found in the text, it is passed to Module 3, where the learning set is enriched.
7. Text, without any slang or suspicious word is derived for further use.
8. Stop.

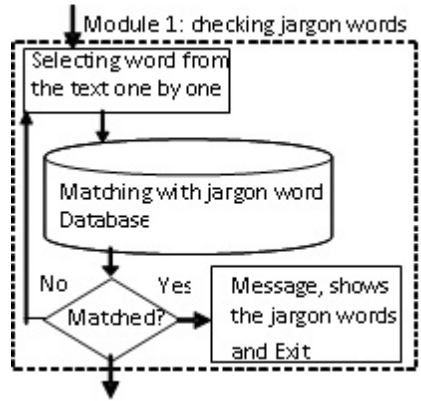

Figure 2. Modular representation of the slang word detection module

**Module 1: Algorithm 2:** This algorithm detects the barely appeared jargon words using supervised learning. A hand tagged Database of slang words is considered (Figure 2) as a learning set. The Time Complexity of the algorithm is $O(n^2)$. Complexity of picking up the n number of words from the text is of $O(n)$ and checking the each word with the entries of the Database is $O(n)$.

Input: Text, containing only meaningful words.
Output: Text, without slang words.

1. Repeat steps 2, 3 and 4 for each word of the text.
2. A word from the text is taken.
3. The word is matched with each entry of the slang word Database.
4. If the word is matched with any entry, the algorithm stops proceeding and Exit.
   Else: Goto step1 loop.
5. Stop.

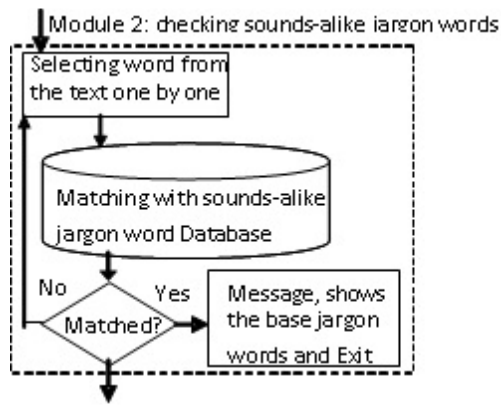

Figure 3. Modular representation of the sounds-alike slang word detection module

**Module 2: Algorithm 3:** This algorithm detects the sounds-alike abbreviated forms of slang words using a Database (Figure 3) of related entries. The Time Complexity of the algorithm is $O(n^2)$, derived in the same way as in Module 1: Algorithm 2.

Input: Text, free from completely matched slang words.
Output: Text, without slang words.

    1. Repeat steps 2, 3 and 4 for each word of the text.
    2. A word is taken from the text.
    3. The word is matched with each entry of the sounds-alike slang word Database.
    4. If the word is matched with any entry,
        A message displays the actual slang word, which must be changed to proceed and Exit.
   Else: Goto step 1 loop.

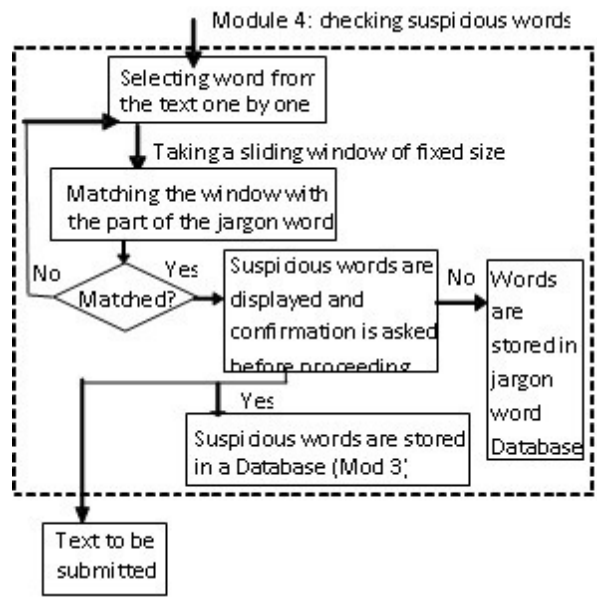

Figure 4. Modular representation of the suspicious word detection module

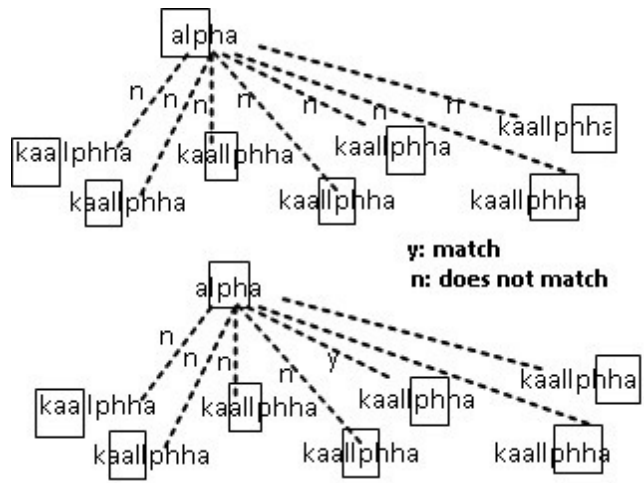

Figure 5. Modular representation of the sliding window mechanism

**Module 4: Algorithm 5:** This algorithm finds the words, which are suspicious in nature to be a jargon word (Figure 4) using sliding window mechanism. If any word partially matches with a jargon word, that word is treated as a suspicious word. The Time Complexity of the algorithm is $O(n^2)$ due to the nesting of operations at Step 1 and Step 4.

> Input: Text, without slang word.
> Output: a) Text, which can be used for proceeding and
>         b) List of Suspicious words.
>
> 1. Repeat step 2, 3 and 4 for each word of the input text.
> 2. A word is taken.
> 3. A character window (Figure 5) of some fixed length is taken and set at the beginning of the word.
> 4. Repeat step 5 till the window lies within the word length.
> 5. If the window is matched with any part of the slang words,
>       A message, mentioning the suspicious word is displayed and waits for confirmation to proceed.
>             If the user confirms the suspicious word as a slang word,
>                 the word is stored in slang word Database directly
>                 and exit.
>       Else,
>             The suspicious word is passed to module 3 for enriching the learning set.
>
>    Else,
>             The window is shifted right by one character.
> 6. The text is displayed.
> 7. Stop.

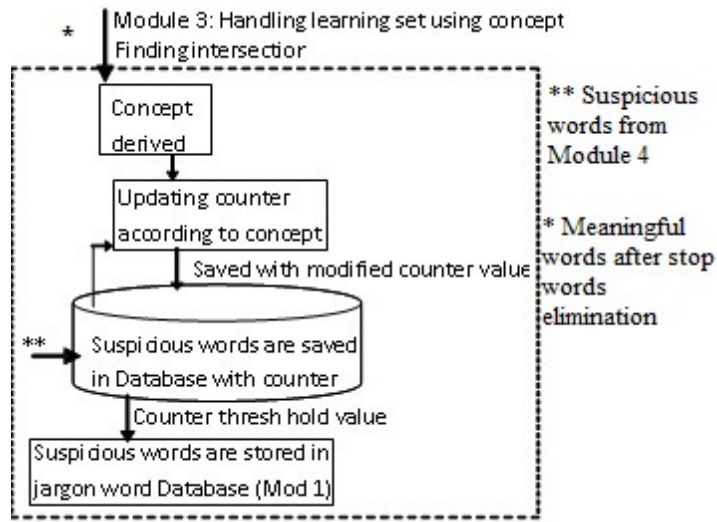

Figure 6. Modular representation of the learning mechanism

**Module 3: Algorithm 4:** This algorithm evaluates the probability of a suspicious word to be a jargon word by the help of a learning set (Figure 6), which increases the efficiency of the jargon word detection procedure. The maximum Time Complexity of the algorithm is $O(n^2)$, which is evaluated at step 1.

> Input: a) Text, containing the only meaningful words from the input text and
> b) Suspicious words from module 4.
> Output: Slang words would be stored in slang word Database (Module 1).
>
> 1. Intersection (Figure 6) is performed between the words of the input text and the different synsets.
> 2. The concept is derived from the highest value of intersection.
> 3. Some predefined weight is assigned to the derived concept.
> 4. Suspicious words from Module 4 are stored in a Database with some counter.
> 5. Repeat steps 6, 7 and 8 for each suspicious word.
> 6. The counter value of a suspicious word is taken.
> 7. The counter value is updated by the weight, assigned for the concept.
> 8. If the counter value of a suspicious word crosses some thresh hold value, the suspicious word is treated as a slang word and is stored in slang word Database (Module 1) for further decision making.
> 9. Stop.

## 5. OUTPUT AND DISCUSSION

The experiment was started with a data set of commonly used slang words (Figure 7) and a table, consists of different *synset*s and the related *concept*s (Figure 8). A weight is assigned to each concept from real life experience ("assgval" in Figure 8), which is used further for learning mechanism. For experiment few Greek words are considered as slang words

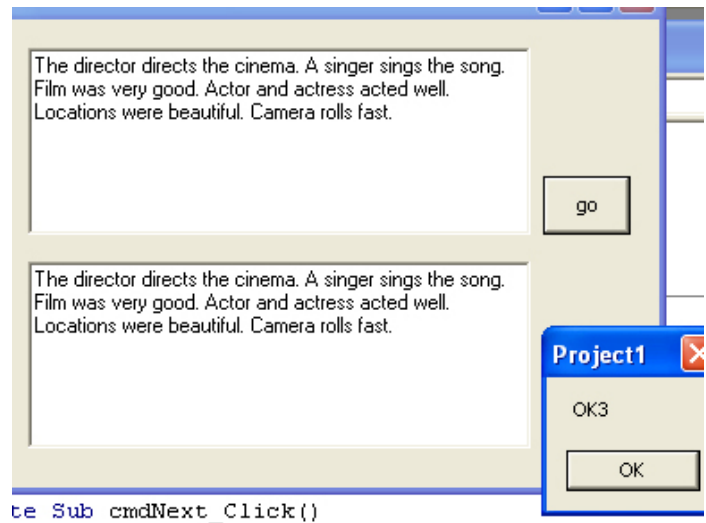

Figure 7. The table, containing few slang words

Figure 8. The table of different synsets and related concepts

In Figure 9, the input text proceeds further as there is no jargon word in the input text.

Figure 9. The input text proceeds further as there is no slang word in the text.

Figure 10 depicts the existence of few slang words in an input text.

International Journal of Artificial Intelligence & Applications (IJAIA), Vol. 4, No. 5, September 2013

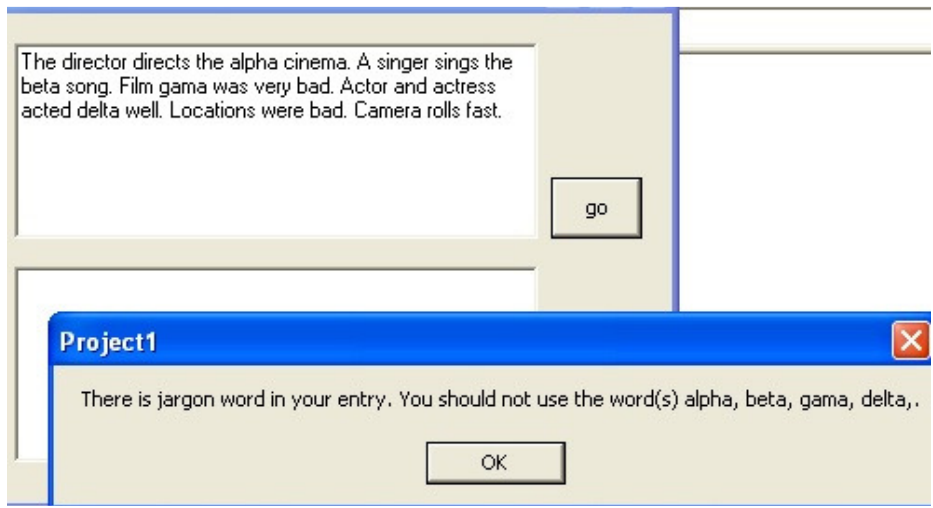

Figure 10. Few jargon words are detected by supervised learning

Figure 11 depicts the detection of sounds-alike abbreviated forms of different slang words in a text.

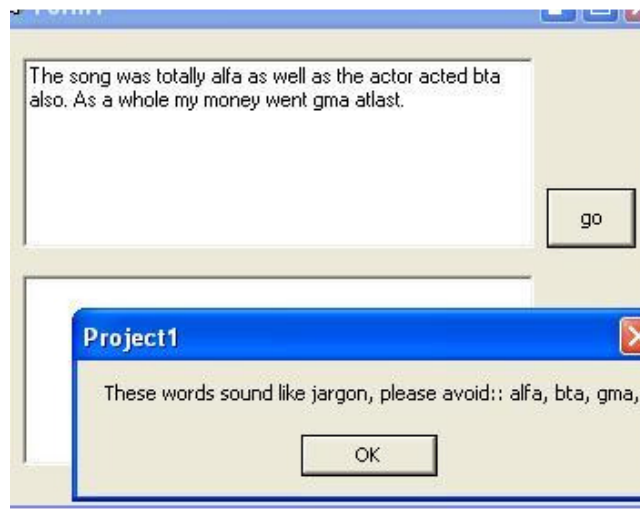

Figure 11. Few sounds-alike abbreviated forms of slang words are detected in the input text

Figure 12 depicts the detection suspicious words in an input text.

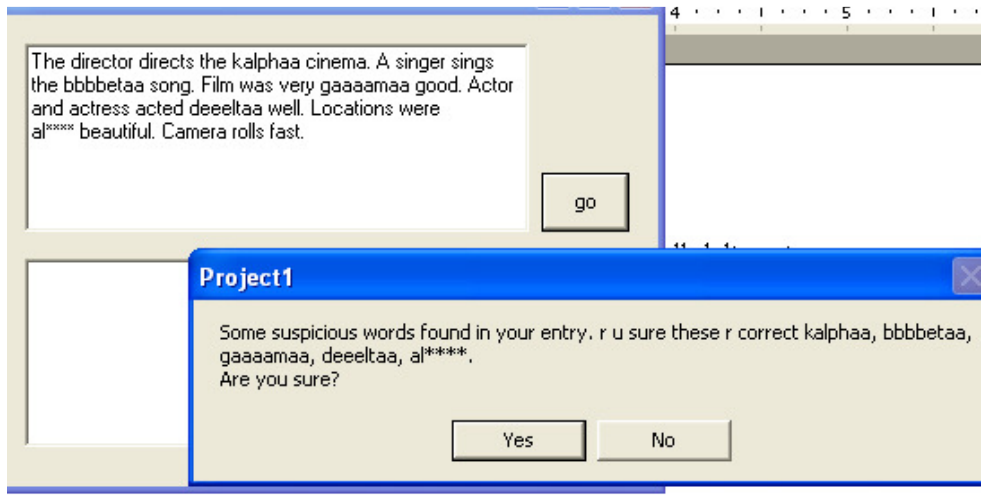

Figure 12. The algorithm alerts the user for verification of the detected suspicious words

If the user accepts a suspicious words as a jargon word, the suspicious words are stored in the jargon word Database for further decision making. Otherwise, the suspicious words are stored in a temporary Database with some counter ("JCount" in Figure 13) for further verification. Counter denotes the frequency of occurrence of that suspicious word in different contexts.

Figure 13. The suspicious words are stored in a temporary Database with a counter value.

Now, the learning ability of the algorithm is discussed with example.
Figure 8 depicts concept analysis of a discussion. The different synsets and associated concepts are formulated for experiment. From our real life observation, it is obvious that the probability of a suspicious word, to be a slang word, is changed according to different contexts. In the "assgval" column, these different values are stored from real life experience.

In figure 12, the context of the discussion is derived as "Movie". So, the suspicious words were stored in to the temporary Database (Figure 13), with the pre-assigned weight 10 (column "JValue" in Figure 13).

Then another text is considered, which is related to "Sports" and the suspicious words are stored into the temporary Database with pre-assigned weight 7 (Figure 14).

International Journal of Artificial Intelligence & Applications (IJAIA), Vol. 4, No. 5, September 2013

If any suspicious word is used repeatedly in different contexts, the associated "JValue" would be increased (Figure 14). If the "JValue" of any particular suspicious word crosses some threshold value, that suspicious word would be treated as a slang word and that would be stored in the jargon word Database (Figure 15) as those words can participate in decision making from the next time. Experimentally, threshold value is taken 50 in this case.

| JID | JWord | JCount | JValue |
|---|---|---|---|
| 71 | eeeeepsilon | 1 | 7 |
| 72 | theeeeta | 1 | 7 |
| 73 | lambddddaaa | 1 | 7 |
| 74 | uuupppsilon | 1 | 7 |
| 79 | kalphaa | 4 | 40 |
| 80 | bbbbetaa | 4 | 40 |
| 81 | deeeltaa | 4 | 40 |
| 82 | al**** | 4 | 40 |
| )er) | | | 0 |

Figure 14. "JValue" of a suspicious word is increased, as it is used repeatedly

| | Slng |
|---|---|
| 10 | alpha |
| 11 | beta |
| 12 | gamma |
| 13 | delta |
| 21 | ********** |
| 37 | epsilon |
| 40 | lambda |
| 41 | upsilon |
| 59 | kalphaa |
| 60 | bbbbetaa |
| 61 | deeeltaa |
| 62 | al**** |
| imber) | |

Figure 15. A suspicious is treated as a slang word, as its 'JValue' crosses thresh hold value

In this way, as much as the algorithm is trained by different input texts of different contexts, the jargon word Database becomes stronger. The "assgval" and all other user assigned values are considered from the real life observation and those values may vary situation wise

## 6. CONCLUSION AND FUTURE WORK

This algorithm detects the barely appeared jargon words, different abbreviated forms of jargon words and suspicious words from an e-text, which is used in any open medium. If the learning mechanism is handled with proper synset and probability analysis, the algorithm can solve a big problem of recent day.

But, in some cases like medical field, judicial system etc. few words are used, which are considered as jargon words in our normal life. These situations should be considered as exceptional cases

**Authors**

Alok Ranjan Pal has been working as an a Assistant Professor in Computer Science and Engineering Department of College of Engineering and Management, Kolaghat since 2006. He has completed his Bachelor's and Master's degree under WBUT. Now, he is working on Natural Language Processing. 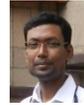

Dr. Diganta Saha is an Associate Professor in Department of Computer Science & Engineering, Jadavpur University. His field of specialization is Machine Translation/ Natural Language Processing/ Mobile Computing/ Pattern Classification.